\documentclass[11pt]{article}

\usepackage[margin=1in]{geometry}
\usepackage[T1]{fontenc}
\usepackage{amsmath, amssymb}
\usepackage{booktabs}
\usepackage{graphicx}
\usepackage{xcolor}
\usepackage{hyperref}
\usepackage[numbers,sort&compress]{natbib}

\DeclareMathOperator{\softmax}{softmax}
\DeclareMathOperator{\silu}{SiLU}
\newcommand{\R}{\mathbb{R}}
\newcommand{\dm}{d_{\text{model}}}
\newcommand{\dffn}{d_{\text{ffn}}}
\newcommand{\de}{d_{e}}

\title{Dense Mixture-of-Experts as a Reparameterized Wide FFN:\\
A Granularity Sweep at Fixed Compute}
\author{
Vu Quang Hoang$^{1,2}$, Nghia Hieu Nguyen$^{1,2}$
\\[1em]
\normalsize
$^{1}$ University of Information Technology\\
$^{2}$ Vietnam National University, Ho Chi Minh City, Viet Nam\\[0.3em]
\texttt{25520611@gm.uit.edu.vn, nghiangh@uit.edu.vn}
}
\date{October 1, 2026}

\begin{document}
\maketitle

\begin{abstract}
Sparse Mixture-of-Experts (MoE) models scale parameter count beyond a token's active compute by routing each token to $K$ experts drawn from a much larger pool of $N \gg K$. This raises a basic question: does MoE's benefit come from the \emph{dynamic, learned combination} of experts, or specifically from the \emph{combinatorial capacity} of selecting from a large pool? We isolate the first factor with a fully dense analogue: a fixed set of $K$ complete SwiGLU experts, all active on every token, combined by a per-token softmax gate, with no larger pool ($N=K$) and no discretization, at fixed total FFN width, so the dense baseline is the $K=1$ case of the same family. Validation loss is non-monotonic in $K$: $K=2$ finishes \emph{below} the dense baseline ($-0.0048$), while $K=4$ and $K=6$ finish above it ($+0.0053$ and $+0.0197$), with the gap to the baseline growing with $K$. All three expert models start behind the baseline early in training; $K=2$ overtakes it between steps 4000 and 4500 and stays ahead for the rest of training, while $K=4$ and $K=6$ never do, and all three gaps plateau by roughly step 9000. Router diagnostics mostly rule out collapse: averaged over layers, per-token gate entropy is moderate ($\approx 0.66$--$0.72$ normalized) and expert usage is fairly balanced ($\geq 0.90$ normalized) for every $K$. One layer is a clear exception: in the $K=4$ model, layer 0's gate is nearly one-hot (91\% of tokens have a gate weight above 0.9, usage $93/1/5/1\%$ across experts), while every other layer in every model stays soft. Pairwise cosine similarity between that model's expert outputs is far higher in layer 0 (mean 0.41) and in the last layer (mean 0.39) than in any layer between them (0.02--0.16), plausibly reflecting each layer's structural position (proximity to the embedding; convergence toward the output head) rather than functional interchangeability: forcing layer 0's gate to uniform at evaluation increases loss by 2.4 nats on a fixed diagnostic subset, showing its collapsed routing is load-bearing, not redundant in any sense that makes it safe to override. We also show that the architecture is exactly a dense SwiGLU of width $K\de$ whose hidden units are scaled by token-dependent, simplex-constrained group weights, so the dense baseline is contained in its function class. These results are consistent with a granularity sweet spot for a softly gated, always-active FFN at fixed total width, with $K=2$ best among the values tested and larger $K$ costing more than it buys; we report this from single runs per configuration.
\end{abstract}

\section{Introduction}
Sparse Mixture-of-Experts architectures have become a standard way to decouple a language model's total parameter count from its per-token compute. A router selects the top-$K$ of $N$ experts for each token, with $N$ often an order of magnitude or more larger than $K$~\citep{shazeer2017outrageously,fedus2022switch,deepseekv3}, so only the selected experts' parameters are used. Making this practical requires solving a set of systems and optimization problems: load-balancing auxiliary losses, expert-capacity limits, token dropping under imbalance, and all-to-all communication across expert-parallel devices.

This machinery exists to make routing \emph{discrete} and therefore cheap; evaluating all $N$ experts densely would defeat the purpose. Discreteness is a computational convenience, not obviously a source of quality. The original Mixture-of-Experts formulation~\citep{jacobs1991adaptive} used a fully dense, differentiable gate over all experts, and sparse routing was introduced later to make large expert pools affordable~\citep{shazeer2017outrageously}.

This leaves a question distinct from ``does MoE help'' and from ``does a larger total pool help at fixed active compute'' (both well supported~\citep{wael2026tiny}): \emph{does the dynamic per-token combination mechanism itself provide any benefit, independent of pool size?} We test this with the simplest instance: a fixed, small set of $K$ complete experts, all active on every token, combined by a learned softmax gate. There is no routing, no discretization, and no larger pool ($N=K$). The architecture is matched to a dense SwiGLU transformer in parameters and FLOPs, and it generalizes the baseline: the gate is near-uniform at initialization and can, in principle, learn behavior close to a single FFN. In the worst case we would expect parity.

Validation loss turns out to be non-monotonic in $K$ at fixed total FFN width, with the dense baseline as $K=1$: $K=2$ ends below the baseline, and $K=4,6$ end above it by an increasing margin (Section~\ref{sec:results}). All expert models trail the baseline early in training, and only $K=2$ overtakes it, doing so partway through training and holding a stable lead afterward, while $K=4,6$ never overtake it and instead plateau at a persistent deficit. Router diagnostics mostly rule out collapse: entropy and usage stay moderate to balanced for almost every layer of every model, with one clear exception, a single layer in the $K=4$ model whose gate is nearly one-hot. We also show the architecture is exactly a dense wide FFN with token-dependent, group-wise scaling of hidden units, which contains the baseline as a special case; this frames the $K$ sweep as a granularity study of that one added mechanism.

Our contributions are:
\begin{enumerate}
  \item A controlled, FLOP- and parameter-matched comparison between a dense SwiGLU transformer and a dense transformer with an always-active, dynamically-weighted set of $K$ experts, at small scale.
  \item An exact algebraic equivalence (Section~\ref{sec:equiv}): the model is a dense SwiGLU of width $K\de$ with token-dependent, simplex-constrained group scaling of hidden units, so the dense baseline is the $K=1$ case of the same family.
  \item A methodological caution: an early conclusion that the baseline wins, drawn from single-batch logged training loss during training, did not hold on validation loss at the end of annealing.
  \item A granularity sweep at fixed total FFN width: $K=2$ finishes below the dense baseline in validation loss, while $K=4$ and $K=6$ finish above it by a growing margin; only $K=2$'s loss curve crosses the baseline's during training, while $K=4,6$ never do and instead plateau at a persistent deficit.
  \item Router diagnostics for $K=2,4,6$, layer by layer: gates are mostly soft and experts fairly evenly used, with one clear exception, a single collapsed layer in the $K=4$ model, which a layer-averaged summary would miss.
  \item Practical guidance for small-scale MoE-inspired research, where details of this kind can reverse an apparent result.
\end{enumerate}

\section{Related Work}

\paragraph{Classical dense-gated and sparse MoE.}
The original MoE formulation~\citep{jacobs1991adaptive} uses a differentiable gate over \emph{all} experts, every expert processes every input, and outputs are combined by the gate distribution. This is the direct ancestor of the design studied here. Sparse top-$k$ routing was introduced to make much larger expert pools tractable~\citep{shazeer2017outrageously}; GShard~\citep{lepikhin2020gshard} and Switch Transformer~\citep{fedus2022switch} developed the load-balancing and capacity machinery, and ST-MoE~\citep{zoph2022stmoe} introduced the router z-loss for stability. Fine-grained expert segmentation~\citep{dai2024deepseekmoe,deepseekv3} pushes $N$ far beyond $K$ (DeepSeek-V3 uses 256 routed experts with 8 active plus a shared expert), relying on combinatorial selection to keep narrow experts expressive. We return to the pre-sparse formulation deliberately, as an instrument for isolating what dynamic combination contributes independent of pool size.

\paragraph{Soft routing via token merging.}
Soft MoE~\citep{puigcerver2024soft} avoids discrete routing differently: it forms per-expert ``slots'' as learned weighted averages over \emph{all tokens in the sequence}, processes each slot with its expert, and reconstructs token outputs as weighted combinations over slots. This is a different axis of ``softness'' from ours (mixing tokens before routing, rather than mixing expert outputs per token), and the authors note that it is difficult to apply to autoregressive decoders because slot computation normalizes over the whole sequence; they leave this to future work. Our gate is per-token with no cross-token dependency and is therefore causal by construction.

\paragraph{Soft routing via parameter merging.}
SMEAR~\citep{muqeeth2024smear} constructs a single merged expert whose parameters are the router-weighted average of the experts' parameters, then runs one forward pass through it, keeping compute close to that of a single expert. For a nonlinear expert $f$, $f(u;\sum_i w_i\theta_i) \neq \sum_i w_i f(u;\theta_i)$ in general, so parameter merging and output merging are not interchangeable relaxations. We use output-space combination at $K\times$ the FLOPs of a single expert, matched to a $K\times$-wide dense baseline.

\paragraph{Geometric and theoretical analyses of soft routing.}
The closest empirical setup we found compares dense, top-$k$, and fully-soft routing under matched capacity, but through representational geometry (Jacobian spectra, effective rank) rather than training dynamics or task loss~\citep{liu2026moe_soft_clustering}; its fully-soft configuration is architecturally close to ours. On the theory side, convergence guarantees exist for jointly trained soft-routed MoE with nonlinear routers and experts under weighted-sum aggregation~\citep{liao2026guided}. Whether the expert redundancy we observe (Section~\ref{sec:router}) matches the feature-learning behavior these analyses predict is beyond our scope.

\paragraph{Causality-preserving differentiable routing.}
A differentiable soft top-$k$ relaxation~\citep{zasada2026softmoe} keeps routing per-token (hence causal) while relaxing the hard top-$k$ operator, with a learned per-layer expert budget. (It shares the short name ``SoftMoE'' with, but is unrelated to, the method of Puigcerver et al.) It still targets genuine sparsity, whereas our design is dense over a small fixed $K$.

\paragraph{Granularity, dense training, and FFN splitting.}
Granularity, the ratio of expert width to dense FFN width, was introduced as an explicit hyperparameter for sparse MoE, and the common choice of setting expert size equal to the dense FFN width was found to be suboptimal at almost any compute budget~\citep{ludziejewski2024scaling}. At larger scale, sparse MoE with an activation rate in a suitable range is reported to beat a dense model with the same total parameters, compute and data~\citep{li2026moe_equal}. Other work uses dense computation inside MoE training: DS-MoE trains with all experts active and routes sparsely only at inference~\citep{pan2024dsmoe}, and a lightweight method gives the router a dense gradient while keeping sparse activation~\citep{panda2025default}. On the FFN side, the sum of parallel FFNs with a shared input is equivalent to one wider FFN~\citep{bercovich2025ffnfusion}, which matches our uniform-gate case; MoEfication splits a pretrained FFN into $k$ experts of width $d_{\text{ff}}/k$ to sparsify it~\citep{zhang2022moefication}; and a multi-branch FFN variant in machine translation was reported to slightly hurt performance~\citep{fan2020mat}. None of these studies sweeps the number of groups of an always-active, softly gated FFN at fixed total width.

\paragraph{Positioning.}
Within the searches we ran (dense-gated MoE, soft routing, granularity studies, FFN splitting), we did not find prior work that varies the number of groups $K$ of a softly gated, always-active FFN at fixed total width, with the dense baseline as $K=1$, and analyzes it through an exact equivalence to a wide dense FFN. Prior soft-routing work either targets large $N$ with capacity gains as the goal, targets single-expert FLOPs, or studies geometry.

\section{Method}
\label{sec:method}

\subsection{Dense baseline}
Each block is pre-norm (RMSNorm~\citep{zhang2019rmsnorm}) with rotary position embeddings~\citep{su2021rope} and multi-head causal attention; tokens embeddings are tied to the output head. The baseline FFN is SwiGLU~\citep{shazeer2020glu}:
\begin{equation}
  \mathrm{FFN}(x) = W_{\text{down}}\big(\silu(W_{\text{gate}}x)\odot W_{\text{up}}x\big),
\end{equation}
with $W_{\text{gate}},W_{\text{up}}\in\R^{\dffn\times\dm}$ and $W_{\text{down}}\in\R^{\dm\times\dffn}$.

\subsection{Dense dynamic mixture of experts}
We replace the FFN with $K$ complete SwiGLU experts $E_1,\dots,E_K$, each of hidden width $\de$, and a single linear router $W_r\in\R^{K\times\dm}$:
\begin{align}
  E_k(x) &= W^{(k)}_{\text{down}}\big(\silu(W^{(k)}_{\text{gate}}x)\odot W^{(k)}_{\text{up}}x\big),\\
  g(x) &= \softmax\!\big(s\cdot W_r x\big)\in\Delta^{K-1},\\
  y(x) &= \sum_{k=1}^{K} g_k(x)\,E_k(x),
\end{align}
where $s\in\{1,\,1/\sqrt{\dm}\}$ is a router logit scale (an ablation factor). All $K$ experts are always evaluated, so there is no top-$k$ selection, no capacity factor, no auxiliary balancing loss, and $g(x)$ depends only on the current token, so decoding remains causal and incremental.

\paragraph{Matching.}
We set $K\de=\dffn^{\text{dense}}$. Core FFN parameters are then $3\dm\dffn^{\text{dense}}$ for both models; the router adds $K\dm$ parameters. Table~\ref{tab:params} gives exact configurations and total parameter counts.

\begin{table}[h]
\centering
\begin{tabular}{lcccc}
\toprule
$K$ & $\de$ & Router params & Non-embedding params & Total params (tied embed./head) \\
\midrule
1 (dense) & $4\dm=1536$ & 0     & 28{,}321{,}152 & 47{,}619{,}840 \\
2         & $2\dm=768$  & 768   & 28{,}330{,}368 & 47{,}629{,}056 \\
4         & $\dm=384$   & 1{,}536 & 28{,}339{,}584 & 47{,}638{,}272 \\
6         & $\tfrac{2}{3}\dm=256$ & 2{,}304 & 28{,}348{,}800 & 47{,}647{,}488 \\
\bottomrule
\end{tabular}
\caption{Exact parameter counts. 12 layers, $\dm=384$, 6 heads, head dim 64, vocab 50{,}257, embeddings tied with the output head. The router adds at most 2{,}304 parameters total (0.005\% of the model), which we treat as negligible for matching purposes.}
\label{tab:params}
\end{table}

\subsection{Relation to a wide dense FFN}
\label{sec:equiv}
Let $h_k(x)=\silu(W^{(k)}_{\text{gate}}x)\odot W^{(k)}_{\text{up}}x\in\R^{\de}$, $h(x)=[h_1(x);\dots;h_K(x)]\in\R^{K\de}$, and $D=[W^{(1)}_{\text{down}},\dots,W^{(K)}_{\text{down}}]\in\R^{\dm\times K\de}$. Stacking the gate and up matrices, $h(x)$ is exactly the hidden activation of a dense SwiGLU of width $K\de$, and
\begin{equation}
  y(x)=D\,\mathrm{diag}\big(g(x)\otimes\mathbf{1}_{\de}\big)\,h(x).
\end{equation}
That is, the model is a dense SwiGLU of width $K\de$ whose hidden units are multiplied, before the down projection, by token-dependent scalars that are shared within each group of $\de$ units and constrained to the simplex. Two consequences follow. (i) At a uniform gate, $g=\mathbf{1}/K$, the model equals a dense SwiGLU of width $K\de$ with down projection $D/K$; since $D$ is free, the function class contains that of the baseline, so a deficit is not a lack of expressiveness. (ii) The only extra flexibility relative to the wide dense FFN is the group-wise, token-dependent scaling, and whether it helps is an empirical question. The simplex constraint also fixes the average scale at $1/K$, so matching a dense FFN's output scale requires larger down-projection weights, a reparameterization that can change optimization even when the function class is the same.

\subsection{Initialization and router variants}
\label{sec:init}
Dense weights use PyTorch's default fan-in scaling, i.e.\ $U(-b,b)$ with $b=1/\sqrt{\text{fan-in}}$. For the expert model we compare:
\begin{itemize}
  \item \textbf{Scaled init:} $b_{\text{gate/up}}=1/\sqrt{\dm K}$ and $b_{\text{down}}=1/\sqrt{\de K}$ (an extra $1/\sqrt{K}$ intended to preserve variance under summation).
  \item \textbf{Fan-in init:} $b_{\text{gate/up}}=1/\sqrt{\dm}$ and $b_{\text{down}}=1/\sqrt{\de}$, identical per-expert scaling to the baseline.
\end{itemize}
The router is initialized $\mathcal{N}(0,0.005^2)$ so $g(x)\approx 1/K$ at step 0. Because $g$ sums to one, the combination already shrinks output variance by roughly $\sum_k g_k^2\approx 1/K$ relative to a single expert, so the extra $1/\sqrt K$ factor compounds that shrinkage. Table~\ref{tab:initaudit} shows a step-0 audit (Gaussian inputs, $\dm=384$, $K=3$, $\de=384$ vs.\ dense $\dffn=1152$).

\begin{table}[h]
\centering
\begin{tabular}{lcc}
\toprule
FFN block at initialization & Output std & Variance / baseline \\
\midrule
Dense SwiGLU ($\dffn=3\dm$)           & 0.1056 & 1.00 \\
Experts, scaled init                   & 0.0111 & 0.011 \\
Experts, fan-in init                   & 0.0610 & 0.334 \\
\bottomrule
\end{tabular}
\caption{Step-0 output scale of the FFN block (numerical simulation of the initialization scheme with i.i.d.\ Gaussian inputs). The fan-in variant's residual ratio $\approx 1/K$ matches the softmax-averaging prediction; the scaled variant starts roughly an order of magnitude quieter in standard deviation.}
\label{tab:initaudit}
\end{table}

\section{Experimental Setup}
\label{sec:setup}
\paragraph{Models.}
All models are 12-layer decoders with $\dm=384$, 6 attention heads (head dim 64), RMSNorm, RoPE and tied embeddings, using the 50{,}257-token GPT-2 BPE vocabulary; total parameter counts are given in Table~\ref{tab:params}. They differ only in the FFN. Every FFN has total hidden width $4\dm=1536$: the dense baseline ($K{=}1$, $\dffn=1536$) and expert models with $K=2$ ($\de=768$), $K=4$ ($\de=384$) and $K=6$ ($\de=256$). We report this set in place of an earlier $K\in\{3,8\}$ run, which used inconsistent settings across configurations and is superseded here; all results in this paper use one fixed init/router variant (fan-in init, no router logit scale) across every $K$.

\paragraph{Training.}
Sequences of 1024 tokens are packed from concatenated documents (no padding; loss and attention use only a causal mask), tokenized with the GPT-2 BPE vocabulary (50{,}257 tokens). Each optimizer step uses a micro-batch of 16 sequences with 4 gradient-accumulation steps, giving an effective batch of 64 sequences (65{,}536 tokens) per step; batches are not shuffled, so all models see identical data in identical order. The schedule has 500 warmup steps, in which the learning rate grows exponentially from $8\times10^{-5}$ by a factor of 1.0025 per step to a peak of $\approx 2.8\times10^{-4}$, followed by 20{,}000 steps of cosine annealing to $10^{-5}$. Training covers 1{,}343{,}488{,}000 tokens, about 1.03 passes over the $\sim$1.3B-token training subset (so close to a single epoch). Training data is a subsample of FineWeb-10B~\citep{penedo2024fineweb}; validation uses the full FineWeb validation split. Throughout this paper, ``step'' refers to the optimizer step count \emph{after} the 500-step warmup (i.e.\ step $s$ in our tables and figures is the $(s{+}500)$-th optimizer step from initialization); token counts were logged from true step 0 and are unaffected by this offset. Optimization uses AdamW~\citep{loshchilov2019adamw} with weight decay 0.01 (PyTorch defaults otherwise: $\beta_1{=}0.9$, $\beta_2{=}0.999$, $\epsilon{=}10^{-8}$), mixed fp16 precision, on 2$\times$NVIDIA T4 GPUs. No random seed was fixed, so each of the four runs used an independent, uncontrolled initialization and data-loader state; this is part of why we treat the final-loss gaps in Section~\ref{sec:results} as suggestive pending a noise estimate (Section~\ref{sec:setup}, Limitations).

\paragraph{Compared models.}
(i) Dense SwiGLU baseline; (ii) dense dynamic MoE (this work) with $K\in\{2,4,6\}$, all at fixed total FFN width $4\dm$ (Table~\ref{tab:params}).

\paragraph{Metrics.}
We compare models by validation loss, logged every 500 steps on the full FineWeb validation split, disjoint from training. Logged training loss is the single-batch loss at the evaluation step and is too noisy to compare across runs, so we do not use it for comparison (Section~\ref{sec:results} discusses a case where it was misleading). Each configuration was run once, without seed-based replication (Limitations).

\section{Results}
\label{sec:results}

\subsection{Dense dynamic MoE vs.\ dense baseline}
\begin{figure}[h]
\centering
\includegraphics[width=0.95\linewidth]{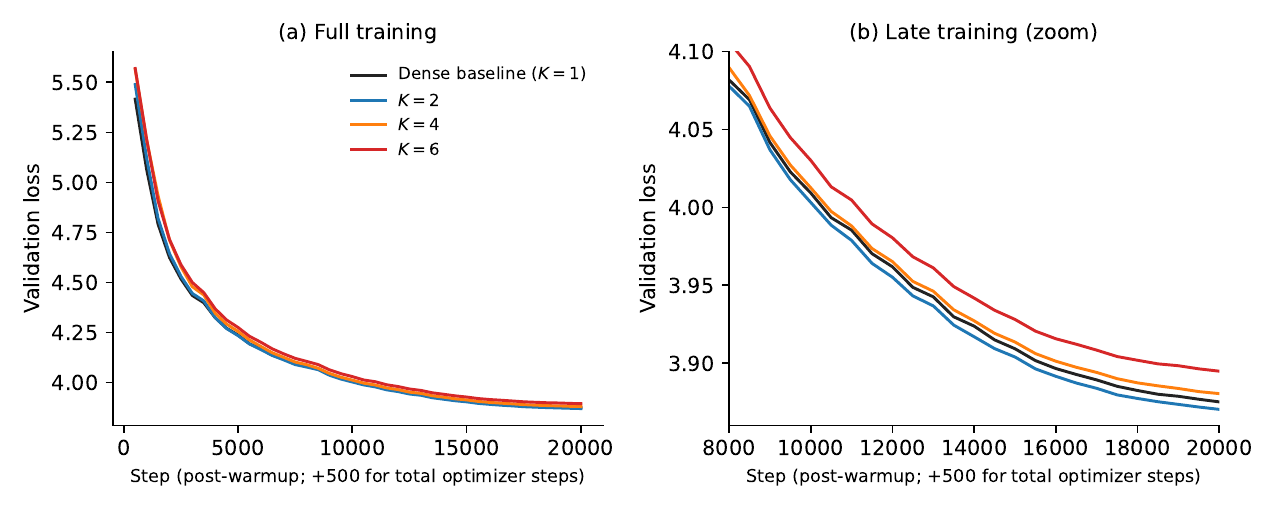}
\caption{Validation loss over 20{,}000 steps for the dense baseline ($K{=}1$) and $K=2,4,6$. (a) Full training. (b) Zoom on steps 8{,}000--20{,}000, where all four curves have entered a slowly-decaying plateau regime.}
\label{fig:curves}
\end{figure}

\begin{figure}[h]
\centering
\includegraphics[width=0.72\linewidth]{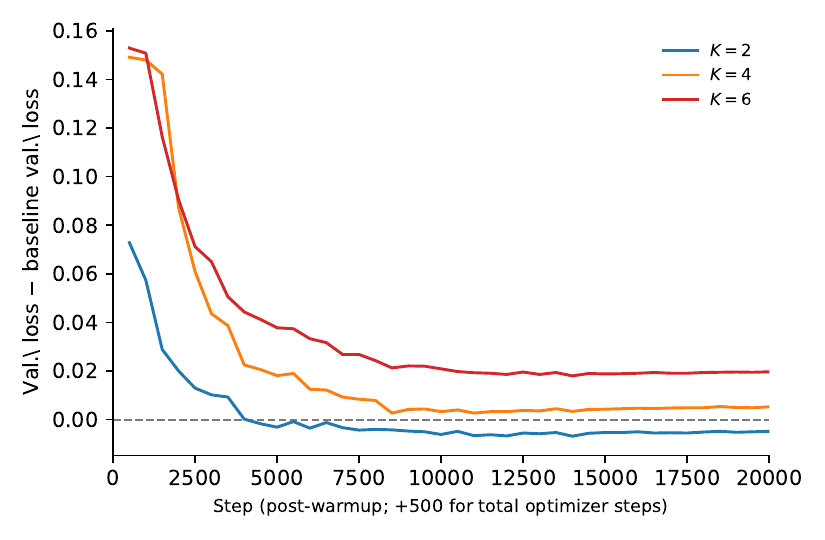}
\caption{Validation loss minus the dense baseline's validation loss, at the same step, for $K=2,4,6$. All three start positive (worse than baseline). $K=2$ crosses zero between steps 4000 and 4500 and stays negative afterward; $K=4$ and $K=6$ never cross zero and plateau at a positive gap that grows with $K$.}
\label{fig:gap}
\end{figure}

\begin{table}[h]
\centering
\begin{tabular}{lccc}
\toprule
Configuration & $\de$ & Val.\ loss (step 20{,}000) & $\Delta$Val vs.\ dense \\
\midrule
Dense SwiGLU ($K{=}1$)      & $4\dm$          & 3.8752 & --- \\
$K{=}2$                      & $2\dm$          & 3.8704 & $-0.0048$ \\
$K{=}4$                      & $\dm$           & 3.8805 & $+0.0053$ \\
$K{=}6$                      & $\tfrac{2}{3}\dm$ & 3.8949 & $+0.0197$ \\
\bottomrule
\end{tabular}
\caption{Validation loss at step 20{,}000. $\Delta$ is expert model minus dense baseline, so negative means the expert model is better. One run per configuration; all use the same init and router variant; parameter counts in Table~\ref{tab:params}.}
\label{tab:main}
\end{table}

Validation loss is non-monotonic in $K$: $K{=}2$ finishes 0.0048 below the dense baseline, while $K{=}4$ and $K{=}6$ finish 0.0053 and 0.0197 above it. Figure~\ref{fig:curves} shows all three expert models start behind the baseline early in training (by 0.07--0.15 at step 500) and close most of that gap within the first few thousand steps. Figure~\ref{fig:gap} shows what happens after: $K{=}2$'s gap crosses zero between step 4000 and 4500 and stays negative (better than baseline) for the remaining 15{,}500 steps, settling around $-0.005$ to $-0.007$ from step 9000 onward. $K{=}4$ and $K{=}6$ never cross zero; their gaps shrink sharply early, then plateau at roughly $+0.004$ and $+0.019$ respectively from step 9000 onward and stay essentially flat for the rest of training. The three curves therefore do not merely differ at the final step: they reach visibly different, stable plateaus well before training ends, which we read as evidence that the differences reflect the architecture rather than still-converging noise.

\subsection{An early, confounded observation at smaller scale}
In earlier experiments at $\sim$5M parameters, an expert model using the scaled init and the $1/\sqrt{\dm}$ router scale from Section~\ref{sec:init} beat a dense baseline by about 0.03 loss, with the gap opening near a token-to-parameter ratio of roughly $20{:}1$ \citep{hoffmann2022chinchilla}. Given that both the scaled init and the router scale are confounds we identify and remove in the runs reported here (Section~\ref{sec:init}), we do not treat that number as informative on its own; we mention it only because the $K{=}2$ result in Section~\ref{sec:results}, obtained without either confound, independently shows the dense baseline being beaten, which is some indication the earlier observation was not purely a confound artifact.

\subsection{Reading the granularity sweep}
\label{sec:granularity}
Table~\ref{tab:main} and Figures~\ref{fig:curves}--\ref{fig:gap} give a single, consistent picture: at fixed total FFN width, going from $K{=}1$ (dense) to $K{=}2$ helps slightly, and going further to $K{=}4$ or $K{=}6$ costs increasingly more. This is not the pattern fine-grained sparse MoE predicts for narrower experts. In fine-grained sparse MoE, narrowing experts while increasing both total pool size $N$ and active count $K$ is beneficial because different tokens can draw on different subsets of a large pool; narrowness is compensated by combinatorial selection~\citep{dai2024deepseekmoe,ludziejewski2024scaling}. Our design has no such pool ($N=K$ always), so narrowing experts here only removes per-branch capacity without adding any compensating selection benefit, and $K{=}4,6$ degrade accordingly. This leaves open why $K{=}2$ specifically improves on the baseline rather than merely degrading least; Section~\ref{sec:equiv}'s equivalence and the router diagnostics below (Section~\ref{sec:router}) are our attempt to explain that specifically, rather than only the ordering.

\subsection{Initialization and router variant}
\label{sec:dynamics}
All results in this paper use one fixed variant throughout: fan-in expert init (Section~\ref{sec:init}) and no router logit scale. During earlier development, removing the $1/\sqrt{\dm}$ router scale and switching from scaled to fan-in expert init both visibly lowered loss faster over the first few hundred steps; those observations came from partially trained runs under a cosine schedule and are not re-verified as a controlled ablation on completed runs.

\subsection{Router diagnostics}
\label{sec:router}
We measured the router at step 20{,}000 on a fixed held-out batch, per layer. We report normalized per-token gate entropy $H/\ln K$ (1 is uniform, 0 is one-hot), the mean maximum gate weight, the fraction of tokens whose maximum gate weight exceeds 0.9, and the normalized entropy of the mean gate (expert usage; 1 means all experts equally used).

\begin{table}[h]
\centering
\begin{tabular}{lcccc}
\toprule
$K$ & Per-token entropy & Mean max weight (uniform) & Frac.\ $>0.9$ & Usage entropy \\
\midrule
2 & 0.719 & 0.762 (0.500) & 0.189 & 0.948 \\
4 & 0.668 & 0.622 (0.250) & 0.097 & 0.901 \\
6 & 0.713 & 0.517 (0.167) & 0.015 & 0.941 \\
\bottomrule
\end{tabular}
\caption{Router statistics at step 20{,}000, averaged over layers, for the models in Table~\ref{tab:main}.}
\label{tab:router}
\end{table}

Averaged over layers, no model looks collapsed: usage entropy is at least 0.90, and per-token entropy is well above zero for all three (0.67--0.72), with mean maximum gate weight clearly above uniform in every case (e.g.\ 0.762 vs.\ 0.500 for $K=2$), so the gate is doing real, non-trivial, token-dependent work without degenerating.

Layer averages hide a real exception. The per-layer values behind Table~\ref{tab:router} show that in the $K{=}4$ model, layer 0 is qualitatively different from every other layer in every model we measured: its per-token entropy is 0.185 (vs.\ 0.48--0.85 elsewhere), 90.9\% of tokens have a maximum gate weight above 0.9 (vs.\ 0--40\% elsewhere), and expert usage is $93/1/5/1\%$ (vs.\ roughly balanced elsewhere, usage entropy 0.221 for this layer against 0.87--1.00 for every other layer in every model). This one layer has, in effect, learned close to hard top-1 routing on its own, with no discretization mechanism forcing it to. Every other layer in the $K{=}2$, $K{=}4$, and $K{=}6$ models stays soft (per-token entropy $\geq 0.48$, usage entropy $\geq 0.87$). We do not have an explanation for why this happens in exactly one layer of one configuration. This means our claim that ``no router collapses'' should be read as a layer-averaged, mostly-true statement rather than a universal one: collapse toward a small number of experts does occur here, locally, without any hard top-$k$ operator or load-balancing loss encouraging it.

These statistics come from the final checkpoint only and do not characterize training-time dynamics; whether layer 0's collapse in the $K{=}4$ model appeared early or late in training is unknown.

\subsubsection{Expert redundancy in the $K{=}4$ model}
\label{sec:cossim}
To test whether the collapsed layer's experts are functionally redundant, we computed pairwise cosine similarity between the four experts' outputs $E_k(x)$ in the $K{=}4$ model, per token on a held-out batch, then averaged the mean and maximum (over the $\binom{4}{2}{=}6$ pairs) per layer.

\begin{table}[h]
\centering
\begin{tabular}{lcc}
\toprule
Layer & Mean pairwise cosine sim.\ & Max pairwise cosine sim.\ \\
\midrule
0  & 0.414 & 0.650 \\
1  & 0.161 & 0.249 \\
2  & 0.085 & 0.187 \\
3  & 0.049 & 0.156 \\
4  & 0.044 & 0.143 \\
5  & 0.040 & 0.125 \\
6  & 0.037 & 0.117 \\
7  & 0.027 & 0.107 \\
8  & 0.021 & 0.091 \\
9  & 0.037 & 0.115 \\
10 & 0.061 & 0.143 \\
11 & 0.388 & 0.496 \\
\midrule
Average & 0.114 & 0.215 \\
\bottomrule
\end{tabular}
\caption{Pairwise cosine similarity between the $K{=}4$ model's expert outputs, by layer, on a held-out batch.}
\label{tab:cossim}
\end{table}

Layers 0 and 11, the first and last, are outliers (Table~\ref{tab:cossim}): mean pairwise cosine similarity is 0.41 and 0.39 there, four to twenty times higher than every layer in between (0.02--0.16). Layer 0 is exactly the layer whose router collapsed (Section~\ref{sec:router}), which suggests an immediate hypothesis: if its experts are relatively redundant, the router's precise weighting may matter little there, making the collapse inexpensive rather than costly. Layer 11 is a natural case to test this against, since it is comparably redundant (mean cosine 0.388 vs.\ layer 0's 0.414) but its router did not collapse (entropy 0.821 vs.\ 0.185, usage entropy 0.983 vs.\ 0.221; Section~\ref{sec:router}). Section~\ref{sec:layer0ablation} tests the hypothesis directly, by ablation rather than by inference from cosine similarity alone, and finds against it.

\subsubsection{Layer 0 is load-bearing, not redundant: a direct ablation}
\label{sec:layer0ablation}
The cosine similarity pattern above raises an immediate question: if layer 0's experts are more mutually aligned than elsewhere, is its router's near-one-hot collapse (Section~\ref{sec:router}) inexpensive, in the sense that overriding it should barely change the model's output? We tested this directly by forcing layer 0's gate to uniform ($g=[\tfrac14,\tfrac14,\tfrac14,\tfrac14]$) at evaluation and leaving every other layer untouched.

\begin{table}[h]
\centering
\begin{tabular}{lcc}
\toprule
Configuration & Val.\ loss & Val.\ ppl.\ \\
\midrule
Normal routing (trained gates)      & 3.7493 & 42.49 \\
Layer 0 forced uniform              & 6.1216 & 455.59 \\
\bottomrule
\end{tabular}
\caption{Uniform-gate ablation on layer 0 of the $K{=}4$ model, evaluated on a smaller, fixed subset of the validation split (CPU-only, for speed; see text). Both rows use this identical subset, so the comparison between them is unaffected by the choice of subset; the absolute values are not directly comparable to Table~\ref{tab:main}'s full-validation-split figures.}
\label{tab:layer0ablation}
\end{table}

This ablation was run on CPU for practicality and therefore evaluated on a smaller, fixed subset of the validation split rather than the full split used for the training-time logging behind Table~\ref{tab:main}; the two available ``normal routing'' figures for $K{=}4$ (3.7493 here, 3.8805 there) differ for that reason and should not be compared directly. What matters for this ablation is the comparison within Table~\ref{tab:layer0ablation}, where both rows share the same fixed subset: the effect is not small, forcing uniformity at this one layer, out of twelve, increases loss by roughly 2.4 nats and perplexity by an order of magnitude on that subset. This rules out the hypothesis, which we raised in Section~\ref{sec:cossim} and retract here, that elevated cosine similarity makes layer 0's collapse cheap because the experts are close to interchangeable. Cosine similarity measures the \emph{direction} of each expert's output; it says nothing about which direction is load-bearing or about magnitude. A mean pairwise cosine similarity of 0.41 reflects a moderate, layer-position-dependent alignment (plausibly because layer 0's input, drawn from the embedding through only one attention block, has had comparatively little opportunity to differentiate across experts, and layer 11's output is pulled toward the narrow subspace the tied output head reads from) -- not near-duplication, and not evidence that any one expert is safe to override. The router's concentration of 93\% of its weight on a single expert at layer 0 is better read as having found a specific, necessary transformation, which uniform blending with three imperfectly-aligned alternatives corrupts rather than approximates. Why layer 11, under a plausibly similar structural pressure toward alignment, did not also collapse remains unexplained.

\section{Discussion}
\label{sec:discussion}
\paragraph{Why is there a sweet spot at $K=2$?}
By Section~\ref{sec:equiv}, the only thing the gate adds over a wide dense FFN is a token-dependent, simplex-constrained scaling of $K$ groups of hidden units; the dense baseline is the $K{=}1$ case where that scaling is trivial (always 1). Two groups with a soft, mostly-balanced gate (Table~\ref{tab:router}) is enough to beat the baseline slightly; four or six groups is not, and gets worse as $K$ grows. One reading is a trade-off: splitting the same total width into more, narrower groups gives the gate more degrees of freedom to reweight, but each group is individually less capable (Section~\ref{sec:granularity}), and past $K{=}2$ the second cost outgrows the first benefit. This is consistent with, but not established by, our data; it predicts that expert outputs should be more similar (redundant) at $K{=}2$ than at $K{=}4,6$, and that a static-weight (non-input-dependent) version of $K{=}2$ should still beat the baseline by less than the full dynamic version, since it lacks the reweighting benefit but keeps the narrower groups. Cosine similarity for the $K{=}4$ model (Section~\ref{sec:cossim}) shows the redundancy pattern is not uniform across depth (concentrated in the first and last layers) and is not, by itself, sufficient to explain the one router collapse we observe. This trade-off account remains a hypothesis rather than an established mechanism.

\paragraph{The $K{=}4$ layer-0 collapse.}
That one layer, in one configuration, converges to near-hard routing without any mechanism designed to produce that (Section~\ref{sec:router}) is itself notable: it shows this architecture \emph{can} rediscover something close to discrete top-1 selection on its own, at least locally, purely from gradient descent on a continuous softmax. We initially hypothesized that layer 0's elevated expert-output cosine similarity (Section~\ref{sec:cossim}) made its collapse inexpensive, since near-duplicate experts would make the router's precise weighting inconsequential. A direct ablation (Section~\ref{sec:layer0ablation}) rules this out: forcing layer 0's gate to uniform costs roughly 2.4 nats of validation loss, an order-of-magnitude increase in perplexity, so its collapsed routing is load-bearing rather than safe to override. The elevated cosine similarity is better read as a structural, position-dependent alignment (Section~\ref{sec:layer0ablation}) than as functional redundancy. Why this collapse happens in layer 0 and not the similarly aligned layer 11 remains unexplained.

\paragraph{What the result implies about MoE.}
Because the expert models have the same FLOPs and parameters as the baseline and at least the same function class (Section~\ref{sec:equiv}), the differences we observe (all within about 0.02 validation loss) cannot come from extra capacity in the sense sparse MoE has it. This is consistent with the sparse-MoE literature's attribution of its benefit to selecting from a larger pool ($N\gg K$)~\citep{dai2024deepseekmoe,ludziejewski2024scaling,li2026moe_equal}, which this design cannot do by construction.

\paragraph{Practical advantages retained.}
The design needs no load-balancing loss, capacity factor, or token dropping, and it is causal and incrementally decodable (unlike token-merging Soft MoE), independent of whether $K{=}2$'s small win holds up under further scrutiny.

\section{Limitations}
Scale is small (28--29M non-embedding, 48M total parameters, Table~\ref{tab:params}) and results come from one dataset. Each configuration was run once, with no random seed fixed, so the four runs are independent single draws rather than seed-paired comparisons, and the $-0.0048$/$+0.0053$/$+0.0197$ gaps in Table~\ref{tab:main} are single-run figures; we read the stable plateaus in Figure~\ref{fig:gap} as evidence they reflect the architecture rather than noise, but this is not established by seed-based replication. The $K{=}4$ layer-0 collapse and its ablation come from a single checkpoint, with no training-time trajectory. The layer-0 ablation used a smaller, fixed validation subset for computational reasons (Section~\ref{sec:layer0ablation}); its within-ablation loss increase, not its absolute values, is what is comparable to the rest of the paper. The trade-off account of the $K{=}2$ sweet spot (Section~\ref{sec:discussion}) is a hypothesis, not an established mechanism. We did not train a sparse MoE model or test $K$ values beyond $\{1,2,4,6\}$. These results are specific to this scale, dataset, and architecture.

\section{Conclusion}
At fixed total FFN width, splitting a dense SwiGLU into a small, always-active, dynamically-weighted set of $K$ experts is not uniformly good or bad: $K{=}2$ slightly beats the dense baseline ($K{=}1$) in validation loss and overtakes it partway through training, while $K{=}4$ and $K{=}6$ fall increasingly behind and never overtake it, with all three gaps stable over the last 11{,}000 of 20{,}000 training steps. Router diagnostics mostly show soft, balanced gates, ruling out collapse as a general explanation, but one layer of the $K{=}4$ model collapses toward near-hard routing on its own, without any mechanism encouraging it to. An exact equivalence to a wide dense FFN with token-dependent group scaling of hidden units (Section~\ref{sec:equiv}) frames this as a granularity study: adding one soft reweighting dimension to a fixed FFN budget can help a little, in moderation, and hurt increasingly once experts are split too finely, with no larger expert pool available to compensate. These results come from single runs per configuration, without seed-based replication, and should be read with that scope in mind.

\bibliographystyle{plainnat}
\bibliography{refs}

\end{document}